\documentclass[a4paper, twocolumn]{article}

\usepackage[american]{babel}
\usepackage{authblk}
\usepackage{microtype}
\usepackage{booktabs}
\usepackage{array}
\usepackage{graphicx}
\usepackage{subfigure}
\usepackage{footnote}
\usepackage{url}
\usepackage{hyperref}
\usepackage{bm}
\usepackage{lmodern}
\usepackage{amsmath}
\usepackage{amsthm}
\usepackage{amsfonts}
\usepackage{tikz}
\usetikzlibrary{decorations.pathreplacing}
\usetikzlibrary{arrows,fit,chains,shapes,calligraphy}
\usetikzlibrary{backgrounds}

\begin{document}

%
\title{\textbf{ESA: Entity Summarization with Attention}}

\author[1,2]{Dongjun Wei\thanks{Equal Contribution. }}
\author[$*$1,2]{Yaxin Liu}
\author[1]{Fuqing Zhu\thanks{Correspondence to Fuqing Zhu.}}
\author[1]{Liangjun Zang}
\author[1]{Wei Zhou}
\author[1]{Jizhong Han}
\author[1]{Songlin Hu}
\affil[1]{Institute of Information Engineering, Chinese Academy of Sciences, China}
\affil[2]{School of Cyber Security, University of Chinese Academy of Sciences, China}
\affil[ ]{\{\url{weidongjun, liuyaxin, zhufuqing, zangliangjun, zhouwei, hanjizhong, husonglin}\}\url{@iie.ac.cn}}

\date{}

\renewcommand*{\Affilfont}{\small}

\maketitle

\begin{abstract}
  Entity summarization aims at creating brief but informative descriptions of entities from knowledge graphs. 
  While previous work mostly focused on traditional techniques such as clustering algorithms and graph models, 
  we ask how to apply deep learning methods into this task.
  In this paper we propose ESA, 
  a neural network with supervised attention mechanisms for entity summarization. 
  Specifically, 
  we calculate attention weights for facts in each entity, 
  and rank facts to generate reliable summaries. 
  We explore techniques to solve difficult learning problems presented by the ESA, 
  and demonstrate the effectiveness of our model in comparison with the state-of-the-art methods. 
  Experimental results show that our model improves the quality of the entity summaries in both F-measure and MAP.
  The source code and outputs are available at \url{https://github.com/WeiDongjunGabriel/ESA}~\footnote{This paper was accepted in EYRE@CIKM'2019}.
\end{abstract}

\section{Introduction}
  Since Knowledge Graph (KG) was first formally defined by Google in 2012, 
  it has been widely applied to various work about Artificial Intelligence (AI). 
  KGs serve for describing real-world entities and the relationship among them. 
  The way to represent databases in KGs to describe entities is usually by RDF triples, 
  the same as data representation on the Web. 
  The Resource Description Framework (RDF) is a data model of Semantic Web, 
  it represents semantic information in the form of \textsl{$<$subject, predicate, object$>$}\cite{DeArteaga2019BiasIB}. 
  With knowledge databases rapidly growing up, 
  the amount of entities and relations in KGs simultaneously rises in an alarming rate. 
  This phenomenon increases the difficulty of extracting or focusing on more representative triples. 
  For quickly comprehending lengthy descriptions in large-scale KGs, 
  summarizing useful information to condense the scale of knowledge databases is an emerging problem to be solved. 
  Entity summarization is a method to extract both brief and informative entities, 
  which has attracted keen interest in recent years, 
  due to the fact that the quality of extracted entities is fundamental to derive subsequent knowledge in various semantic tasks.

  Cheng et al. \cite{Cheng2011RELINRA} proposed RELIN to rank features based on relatedness and informativeness for quick identification of entities, 
  which is adapted according to random surfer model. 
  DIVERSUM \cite{Sydow2010DIVERSUMTD} takes the diverse property of entities into consideration for summarizing tasks in knowledge graphs. 
  FACES \cite{Gunaratna2015FACESDE} makes a proper balance between centrality and diversity of extracted triples through Cobweb algorithm. 
  FACES-E \cite{Gunaratna2016GleaningTF}, 
  proposed by Gunaratna et al., 
  optimizes FACES by considering the effect of literals in entity summarization. 
  CD \cite{Xu2016CDAE} follows the idea of binary quadratic knapsack problem to complete entity summarization. 
  LinkSUM \cite{Thalhammer2016LinkSUMUL}, 
  based on PageRank algorithm to rank triples, 
  rather than utilizing the diversity of properties, 
  LinkSUM focuses more on the objects. 
  ESA introduces deep learning methods into entity summarization task, 
  which employs supervised attention mechanism with BiLSTM to generate representative summaries of entities.
  
Most of them focus on specific aspects of entities, which are insufficient to completely describe various relation among entities. Meanwhile, the data 			supplement technique in this task is limited. For instance, the data supplement method proposed by ES-LDA can only be used for specific facts, which is lack of versatility and probability. In this work, to overcome the defects of the above traditional methods, we ask how to apply deep learning methods into entity summarization task. We propose a model for entity summarization called ESA, which uses supervised attention mechanism with BiLSTM. The ESA allows us to calculate attention weights for facts in each entity, then ranking facts to generate reliable summaries. Experimental results show that our model improves the quality of the entity summaries in both F-measure and MAP.

\section{Task Description}\label{TD}
RDF is an abstract data model, 
and an RDF graph consists of a collection of statements. 
Simple statements generally represent real-world entities, 
which are usually stored as triples. 
Each triple $t$ represents a fact that is in the form of <$subject$, $predicate$, $object$>, 
abbreviated as $<s, p, o>$. 
Since RDF data is encoded by unique identifiers (URIs), 
an entity in RDF graphs can be regarded as a subject with all predicates and corresponding objects to those predicates.

\textbf{Definition 1 (Entity Summarization):} 
Entity Summarization (ES) is a technique to summarize RDF data for creating concise summaries in KGs. 
The subject of each entity provides the core for summarizing entities. 
Therefore, 
the task of ES is defined as extracting a subset from a lengthy feature set of each entity with the respective subject. 
Given an entity $e$ and a positive integer $k$, the output is \textsl{top-k} features of every entity $e$ in the ranking list of $ES\left( e, k\right)$.

\section{Proposed Model}\label{P}
\begin{figure}[htbp]\label{fig:network}
  \includegraphics[scale=0.42]{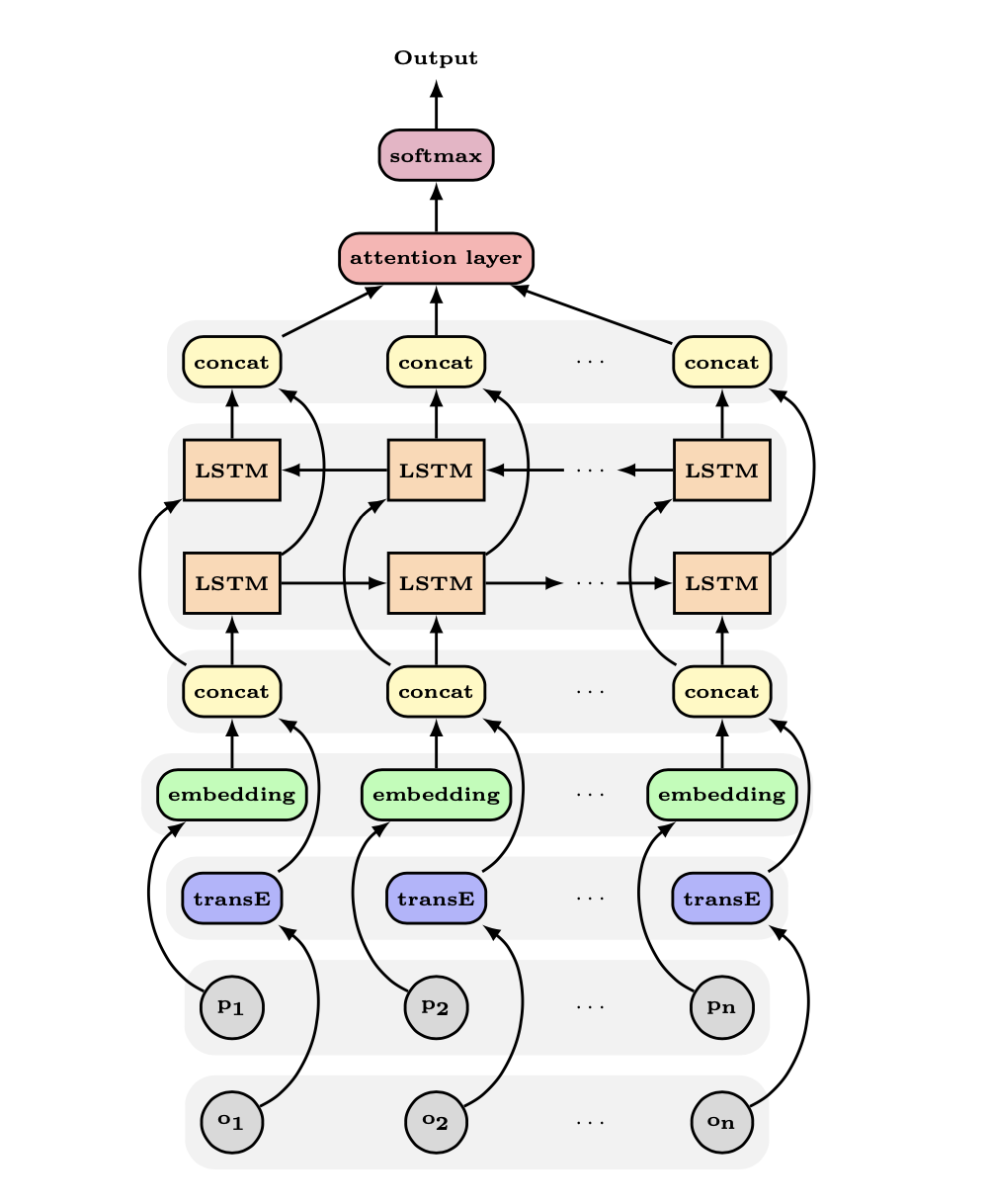}
  \caption{The Architecture of ESA Model}
\end{figure}

We model ES as a ranking task similar to existing work, 
such as RELIN, FACES, and ES-LDA. 
Unlike the traditional approaches to generate entity summaries in KGs, 
the ESA is a neural network model using sequence model, Figure~\ref{fig:network} describes the architecture of the model.

Similar to most sequence models~\cite{Cho2014OnTP}, 
the ESA has an encoder\-decoder structure. 
The encoder is consisted of knowledge representation and BiLSTM, 
it maps an input sequence $\left( t_1, t_2, \ldots, t_n\right)$ of RDF triples from a certain entity to a continuous representation $h= \left(h_1, h_2, \ldots, h_n\right)$. 
The decoder is mainly composed of attention model. 
Given $h$, 
the decoder then uses a supervised attention mechanism generates an output vector $\left( a_1, a_2, \ldots, a_n \right)$ representing attention vector for each entity, 
which is then used as evidence for summarizing entities. 
Higher attention weights are related to more important triples, 
we finally select triples according to \textsl{top-k} highest weights as our entity summaries.
\subsection{Knowledge Representation}
  Entities in large-scale KGs are usually described as RDF triples, 
  while each triple consists of a subject, a predicate, and an object. 
  MPSUM proposed by Wei~\cite{WeiMPUSUM2018} takes the uniqueness of predicates and the importance of objects into consideration for entity summarization.
  The experimental results show that the characteristics of predicates and objects are key factors to select entities.
  In order to make full use of the information contained by RDF triples, 
  we extract predicates and objects from these triples. 
  Let $n$ be the number of triples with the same subject $s$, 
  then two lists respectively based on extracted predicates and objects are $l_1=\left( p_1,p_2, \ldots, p_n \right)$ and $l_2=\left( o_1,o_2, \ldots,o_n\right)$, 
  where $p_i$ and $o_i$ are corresponding predicates and objects from the $i$-th triples. 
  For each entity, 
  we employ different methods to map predicates and objects into continuous vector space respectively~\cite{Lai2016HowTG}.
  In this way, 
  we can balance the difference of occurrence between predicates and objects, 
  which can impact on word embedding of predicates and objects.
\subsubsection*{Predicate Embedding Table}
  We use learned embeddings~\cite{Bengio2000ANP} to convert the predicate input to
  vectors of dimension $d_p$. We randomly initialize embedding vector for each predicate and tune it
  in training phase.
\subsubsection*{Object Embedding Table}
  Unlike generating representation of predicates based on word embedding technique, 
  we use TransE model~\cite{Bordes2013TranslatingEF} to map objects to vectors of dimension $d_o$. 
  We first pretrain transE model based on ESBM benchmark v1.1, 
  and extract the word vectors of objects to construct a lookup table for object vectors. 
  Then we obtain object vectors by looking up the table as input, 
  the object vectors are fixed during training phase.
\subsection*{BiLSTM Network}
    We use BiLSTM network to generate $\bm{h}_s$ and $\bm{h}$ as the input of each general attention layer in multi-head attention mechanism. 
    We can capture contextual information from the forward and backward scan via two sub-networks \cite{Graves2013GeneratingSW}. 
    Specifically, 
    to estimate the importance of $i$-th triple, 
    we employ BiLSTM to extract the information of former triples from $1$ to $i-1$ and later triples from $i+1$ to $n$, 
    where the information respectively propagations forward and backward. 
    In this paper, 
    we denote the $LSTM_L$ and $LSTM_R$ as the forward and backward LSTM model, 
    $x_i$ as the input at the time step $i$ for $LSTM_L$ and $LSTM_R$, 
    and $\bm{h}_{L_i}$ is the output at time step $i$ for the $LSTM_L$, 
    $\bm{h}_{R_i}$for the $LSTM_R$. 
    We encode the input $x_i$ using Bidirectional LSTM as follows:
    \begin{gather}
      \bm{h}_{L_i} = LSTM_L \left( x_i, \bm{h}_{L_{i-1}} \right) \nonumber \\
      \bm{h}_{R_i} = LSTM_R \left( x_i, \bm{h}_{R_{i-1}} \right)
    \end{gather}
    The final output $\bm{h}=\left[ h_1, h_2, \cdots, h_n \right]$, 
    and its component $\bm{h}=\left[ h_1, h_2, \cdots, h_n \right]$ 
    of BiLSTM is calculated by concatenating $\bm{h}_{L_i}$ and $\bm{h}_{R_i}$.

    Moreover, the $\bm{h}_s$ is concatenated by $\bm{h}_s^1$ and $\bm{h}_s^2$. 
    Here, $\bm{h}_s^1$ is the value of hidden state from the final cell of upper LSTM layer, 
    while $\bm{h}_s^2$ is the value of hidden state from the final cell of lower LSTM layer.

\subsection{Supervised Attention}
  Attention Model (AM) is a mainstream neural network in various tasks such as Natural Language Processing~\cite{Young2018RecentTI}~\cite{Vaswani2017AttentionIsAllYouNeed}. 
  For instance, 
  in  machine translation tasks~\cite{Luong2015EffectiveAT}, 
  only certain words in the input sequence may be relevant for predicting the next~\cite{Chaudhari2019AnAS}. 
  AM incorporates this notion by allowing the model to dynamically pay attention to only certain parts of the input 
  that help in performing the task at hand effectively. 
  In entity summarization task, 
  when users observe the facts in each subject, 
  they may pay more attention to certain facts than the rest, 
  which can be modeled based on AM by assigning an attention weight for each fact in the subject. 
  
  In this section, we first introduce the details of constructing
  gold attention vectors and machine attention vectors. 
  Then we describe the loss function and training method in our model,
  which aims at generating machine attention vectors which is similar to the gold attention vectors.
  \subsubsection*{Gold Attention Vectors}
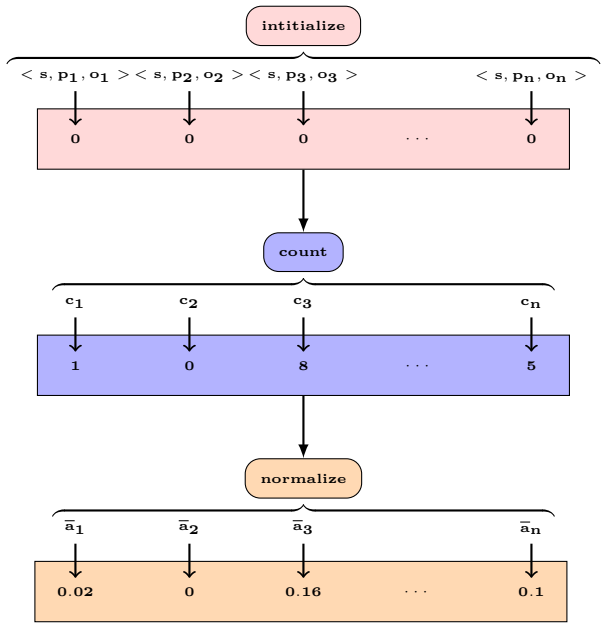
\begin{figure}
\centering
\tikzstyle{init} = [pin edge={to-,thick,black}]
\tikzstyle{background_1} = [
    rectangle,
    draw = black,
    fill=pink!60,
    minimum width=7cm,
    minimum height=0.8cm
]

\tikzstyle{background_2} = [
    rectangle,
    draw = black,
    fill=blue!30,
    minimum width=7cm,
    minimum height=0.8cm
]

\tikzstyle{background_3} = [
    rectangle,
    draw = black,
    fill=orange!30,
    minimum width=7cm,
    minimum height=0.8cm
]

\tikzstyle{normalization} = [
    rectangle,
    draw = black,
    fill = orange!30,
    inner sep=0.2cm,
    rounded corners=2mm
]

\tikzstyle{count} = [
    rectangle,
    draw = black,
    fill = blue!30,
    inner sep=0.2cm,
    rounded corners=2mm
]

\tikzstyle{intitialize} = [
    rectangle,
    draw = black,
    fill = pink!60,
    inner sep=0.2cm,
    rounded corners=2mm
]

\begin{tikzpicture}[>=latex,auto,node distance=1.5cm]
    \tikzstyle{every node}=[font=\tiny]

    \node [intitialize] (intitialize_1) {$\mathbf{intitialize}$};
    
    \node [pin={[init]$\mathbf{<s,p_3,o_3>}$}] (i_3) [below of=intitialize_1] {$\mathbf{0}$};
    \node [pin={[init]$\mathbf{<s,p_2,o_2>}$}] (i_2) [left of=i_3] {$\mathbf{0}$};
    \node [pin={[init]$\mathbf{<s,p_1,o_1>}$}] (i_1) [left of=i_2] {$\mathbf{0}$};
    \node [] (i_4) [right of=i_3] {$\mathbf{\cdots}$};
    \node [pin={[init]$\mathbf{<s,p_n,o_n>}$}] (i_5) [right of=i_4] {$\mathbf{0}$};

    \node [count] (count_1) [below of=i_3] {$\mathbf{count}$};

    \node [pin={[init]$\mathbf{c_3}$}] (a_3) [below of=count_1] {$\mathbf{8}$};
    \node [pin={[init]$\mathbf{c_2}$}] (a_2) [left of=a_3] {$\mathbf{0}$};
    \node [pin={[init]$\mathbf{c_1}$}] (a_1) [left of=a_2] {$\mathbf{1}$};
    \node [] (a_4) [right of=a_3] {$\mathbf{\cdots}$};
    \node [pin={[init]$\mathbf{c_n}$}] (a_5) [right of=a_4] {$\mathbf{5}$};
    
    \node [normalization, below of=a_3] (normal) {$\mathbf{normalize}$};

    \node [pin={[init]$\mathbf{\overline{a}_3}$}] (b_3) [below of= normal] {$\mathbf{0.16}$};
    \node [pin={[init]$\mathbf{\overline{a}_2}$}] (b_2) [left of= b_3] {$\mathbf{0}$};
    \node [pin={[init]$\mathbf{\overline{a}_1}$}] (b_1) [left of= b_2] {$\mathbf{0.02}$};
    \node (b_4) [right of= b_3] {$\cdots$};
    \node [pin={[init]$\mathbf{\overline{a}_n}$}] (b_5) [right of= b_4] {$\mathbf{0.1}$};

    \begin{pgfonlayer}{background}
        \node [background_1,
        fit=(i_1) (i_5),
        ] (back_1) {};
        \node [background_2,
        fit=(a_1) (a_5),
        ] (back_2) {};
        \node [background_3,
        fit=(b_1) (b_5),
        ] (back_3) {};
    \end{pgfonlayer}

    \path[->]
        (back_1) edge[thick] (count_1)
        (back_2) edge[thick] (normal)
        ;
    
    \draw[decorate,decoration={calligraphic brace,amplitude=1.5mm},thick] (-3.9,-.5) -- (3.9,-.5);
    \draw[decorate,decoration={calligraphic brace,amplitude=1.5mm},thick] (-3.3,-3.5) -- (3.3,-3.5);
    \draw[decorate,decoration={calligraphic brace,amplitude=1.5mm},thick] (-3.3,-6.5) -- (3.3,-6.5);
\end{tikzpicture}
\caption{The Process of Constructing of Gold Attention Vectors}
\label{fig:glod-attention}
\end{figure}

  In this work, 
  we use ESBM benchmark v1.1 as our dataset. 
  For each subject we need to summarize,
  ESBM becnchmark v1.1 not only provides the whole RDF triples which is related to this subject, 
  but also provides several sets of \textsl{top-5} and \textsl{top-10} triples selected by different users 
  according to their preference which we can utilize to construct gold attention vectors. 
  We first initialize an attention vector to zero, 
  whose dimension is the number of RDF triples in the subject. 
  Then, 
  we count the frequency of each triple selected by users to update the vector, 
  the $i$-th value $c_i$ in this vector represents the frequency of triple $t_i$. 
  Since ESBM benchmark v1.1, 
  each subject has five sets of \textsl{top-5} and \textsl{top-10} triples selected by five different users, 
  so the frequency of each triple ranges from $0$ to $5$. 
  Figure~\ref{fig:glod-attention} illustrates the details, 
  where $\overline{\alpha}$ is the final gold attention vector after normalization,
  each value in $\overline{\alpha}$ is calculated by the following equation, 
  $\overline{\alpha}_{i}$ denotes the $i$-th value in vector $\overline{\alpha}$:
  \begin{equation}
    \overline{\alpha}_{i} = \frac{c_i}{\Sigma_{i=1}^{n}c_{i}} \,.
  \end{equation}

\subsubsection*{Machine Attention Vectors}
  To generate machine attention vectors with AM, 
  we first obtain the output vectors $h = \left( h_1, h_2, \ldots, h_n\right)$ that the BiLSTM layer produced. 
  Then, 
  the attention layer can automatically learn attention vector $\alpha = \left( \alpha_1, \alpha_2, \ldots, \alpha_n \right)$ based on $h$. 
  We use softmax technique to generate final attention vector $\alpha$:
  \begin{equation}
    \alpha = softmax \left( h_{s}^{T} h\right) \,.
  \end{equation}
  where $h_s$ is concatenated by $h_s^1$ and $h_s^2$, 
  here, $h_s^1$ is the value of hidden state from the final cell of upper LSTM layer, 
  while $h_s^2$ is the value of hidden state from the final cell of lower LSTM layer.
  We rank final attention weight vector $\alpha$, 
  and pick \textsl{top-k} values. 
  Then we obtain the entity summaries based on corresponding \textsl{top-k} values.
\subsubsection*{Training}
  Given the gold attention $\overline{\alpha}$ and the machine attention $\alpha$ produced by our model, 
  we employ cross-entropy loss and define the loss function $L$ of our model as follows:
\begin{equation}
  L\left( \alpha, \overline{\alpha} \right) = CrossEntropy\left( \alpha, \overline{\alpha} \right) \,.
\end{equation}
  Finally, 
  we use backpropagation algorithm to jointly train the whole ESA model.
\section{Experiment}\label{E}
  In this section we introduce the implementation details, 
  and the experimental results on specific datasets. 
  To prove the effectiveness of our model, we take the state-of-the-art approaches to date in entity summarization task for comparison,
   including RELIN, DIVERSUM, CD, FACES-E, FACES, and LinkSUM.
\subsection{Datasets}
  In this work, experiments are conducted based on ESBM Benchmark v1.1 as ground truth. 
  The ESBM benchmark v1.1 consists of $175$ entities 
  including $125$ entities are from DBpedia\footnote{\url{https://wiki.dbpedia.org}} and the rest entities are from LinkedMDB\footnote{\url{http://linkedmdb.org}} datasets. 
  The datasets and ground truth of the entity summarizers can be obtained from~\footnote{\url{http://ws.nju.edu.cn/summarization/esbm/}}. 
  We employ $5$-fold cross validation method for ESBM benchmark v1.1 to construct train sets and test sets.
\subsection{Evaluation Metrics}
  We employ F-measure and MAP as our evaluation metrics. 
  F-measure (so-called F-score or F1-score) is a statistic computed by the harmonic average of the precision and recall, 
  where an F-measure reaches its best at $1$ with perfect precision and recall. 
  MAP (Mean Average Precision) is the mean of AP from multiple datasets, 
  where AP represents average precision for each dataset.
\subsection{Implementation Details}
  We apply word embedding technique to map predicates into continuous space and 
  use pretrained translation vectors with transE for objects. 
  During training phase, 
  the word vectors of predicates are jointly trained 
  while the object vectors are fixed. 
  We use thunlp~\footnote{\url{https://github.com/thunlp/TensorFlow-TransX}} to train the whole ESBM benchmark v1.1.
  We generate gold attention vectors based on ESBM benchmark v1.1, 
  and calculate machine attention vectors based on our model. 
  Finally, 
  we compare our model in terms of \textsl{top-5} and \textsl{top-10} entity summaries with the benchmark results of the entity summarization tools, 
  i.e. RELIN, DIVERSUM, CD, FACES-E, FACES, and LinkSUM, 
  as shown in Table~\ref{tab:F-meausre} and Table~\ref{tab:MAP}.
\subsection{Hyper-parameter Setting}
  Hyper-parameters are tuned on the selected datasets. 
  We set the dimension of predicate embedding to $100$, 
  the dimension of transE to $100$. 
  The learning rate in our model is set to $0.0001$.
\subsection{Experimental Results}
\begin{table*}[htbp]
  \small
  \centering
  \begin{minipage}{\textwidth}
  \begin{center}
  \begin{tabular}{p{1.8cm}p{1.8cm}<{\centering}p{1.8cm}<{\centering}p{1.8cm}<{\centering}p{1.8cm}<{\centering}p{1.8cm}<{\centering}p{1.8cm}<{\centering}}
    \toprule[1.3pt]
    & \multicolumn{2}{c}{\textbf{DBpedia}} 
    & \multicolumn{2}{c}{\textbf{LinkedMDB}} 
    & \multicolumn{2}{c}{\textbf{ALL}} \\
    & k=5 & k=10
    & k=5 & k=10
    & k=5 & k=10 \\
    \midrule[0.6pt]
    \textbf{RELIN}~\cite{Cheng2011RELIN} & 0.242 & 0.455 & 0.203 & 0.258 & 0.231 & 0.399 \\
    \textbf{DIVERSUM}~\cite{Sydow2010DIVERSUMTD} & 0.249 & 0.507 & 0.207 & 0.358 & 0.237 & 0.464 \\
    \textbf{CD}~\cite{DanyunXu2016CD} & 0.287 & 0.517 & 0.211 & 0.328 & 0.252 & 0.455 \\
    \textbf{FACESE}~\cite{Gunaratna2016FACES-E} & 0.280 & 0.485 & 0.313 & 0.393 & 0.289 & 0.461 \\
    \textbf{FACES}~\cite{Gunaratna2015FACES} & 0.270 & 0.428 & 0.169 & 0.263 & 0.241 & 0.381 \\
    \textbf{LinkSUM}~\cite{Thalhammer2016LinkSUMUL} & 0.274 & 0.479 & 0.140 & 0.279 & 0.236 & 0.421 \\
    \textbf{ESA} & \textbf{0.310} & \textbf{0.525} 
    & \textbf{0.320} & \textbf{0.403} & \textbf{0.312} & \textbf{0.491} \\
    \bottomrule[1.3pt]
  \end{tabular}
  \end{center}
  \end{minipage}
  \footnotesize{$^a$ By how much we are better than the best result of all other methods.}
  \caption{Experimental Results on ESBM benchmark v1.1 of F-measure}
  \label{tab:F-meausre} 
\end{table*}

\begin{table*}[htbp]
  \centering  
  \small
  \begin{minipage}{\textwidth}
  \begin{center}
  \begin{tabular}{p{1.8cm}p{1.8cm}<{\centering}p{1.8cm}<{\centering}p{1.8cm}<{\centering}p{1.8cm}<{\centering}p{1.8cm}<{\centering}p{1.8cm}<{\centering}}
    \toprule[1.3pt]
    & \multicolumn{2}{c}{\textbf{DBpedia}} 
    & \multicolumn{2}{c}{\textbf{LinkedMDB}} 
    & \multicolumn{2}{c}{\textbf{ALL}} \\
    & k=5 & k=10
    & k=5 & k=10
    & k=5 & k=10 \\
    \midrule[0.6pt]
    \textbf{RELIN}~\cite{Cheng2011RELIN} & 0.342 & 0.519 & 0.241 & 0.355 & 0.313 & 0.466 \\
    \textbf{DIVERSUM}~\cite{Sydow2010DIVERSUMTD} & 0.310 & 0.499 & 0.266 & 0.390 & 0.298 & 0.468 \\
    \textbf{CD}~\cite{DanyunXu2016CD} & - & - & - & - & - & - \\
    \textbf{FACESE}~\cite{Gunaratna2016FACES-E} & 0.388 & 0.564 & 0.341 & 0.435 & 0.375 & 0.527 \\
    \textbf{FACES}~\cite{Gunaratna2015FACES} & 0.255 & 0.382 & 0.155 & 0.273 & 0.227 & 0.351 \\
    \textbf{LinkSUM}~\cite{Thalhammer2016LinkSUMUL} & 0.242 & 0.271 & 0.141 & 0.279 & 0.213 & 0.345 \\
    \textbf{ESA} & \textbf{0.392} & \textbf{0.582} 
    & \textbf{0.367} & \textbf{0.465} & \textbf{0.386} & \textbf{0.549} \\
    \bottomrule[1.3pt]
  \end{tabular}
  \end{center}
  \end{minipage}
  \footnotesize{$^a$ By how much we are better than the best result of all other methods.}
  \caption{Experimental Results on ESBM benchmark v1.1 of MAP}
  \label{tab:MAP}
\end{table*}
  In this paper, 
  we have carried out several experiments regarding to different metrics based on 
  DBpedia, LinkedMDB, and their combination. 
  The results regarding F-measures are shown in Table~\ref{tab:F-meausre}, 
  and MAPs are shown in Table~\ref{tab:MAP}. 
  ESA achieves better results than all other state-of-the-art approaches 
  not only in each dataset, 
  but also perform best in each metric.
  
  \subsubsection*{F-measure} As shown in Table~\ref{tab:F-meausre}, 
  the best improvement in single dataset is under \textsl{top-5} summaries generated from DBpedia, 
  our model gets the highest F-measure with $0.310$, 
  which excesses the previously best result produced by CD. 
  In terms of DBpedia dataset, 
  the total increase of \textsl{top-5} and \textsl{top-10} summaries is $0.031$. 
  For LinkedMDB dataset, 
  our model obtains the best score both in $k=5$ and $k=10$.
  Meanwhile, we combine two datasets to implement entity summarization, 
  our model has $7.96\%$ and $5.82\%$ increase respectively for the results based on \textsl{top-5} and \textsl{top-10} results.
  
  \subsubsection*{MAP} Our model also gets better scores for MAP metric, 
  as Table~\ref{tab:MAP} shows, where the best increase is $0.030$ represented in LinkedMDB for $k=10$. 
  The improvement of LinkedMDB is more obvious in MAP metric than F-measure, 
  where the total increase is up to $0.056$. 
  
  \subsubsection*{ALL} Combine Table~\ref{tab:F-meausre} and Table~\ref{tab:MAP}, 
  it is evident that our ESA model yields better results both for F-measures and MAPs. 
  It is worth mentioning that our model outperforms all other state-of-art approaches in both F-measure and MAP given by EMBS benchmark v1.1, 
  which can significantly prove the effectiveness of our model.

\section{Conclusion}\label{C}
  In this work, 
  we propose a effective neural network model, 
  called ESA (Entity Summarization with attention).
  Take the human preference into consideration, 
  this model introduces popular notion of attention technique into entity summarization task. 
  Meanwhile, 
  we explore the way to construct gold attention vectors for modelling supervised attention mechanism. 
  The ESA applies extracted predicates and objects as input, 
  in particular, 
  we exploit different but proper knowledge embedding methods respectively for predicates and objects, 
  where the word embedding method is for predicates and TransE is for objects. 
  The final output of ESA is normalized attention weights, 
  which can be used to select representative entities. 
  Our experiments indicate that word embedding technique and graph embedding technique like TransE 
  can be combined together into a single task, 
  which can better represent the fact or knowledge in knowledge graph and provide a more powerful input vectors for neural networks or other models.
  Experimental results show that our work outperforms all other approaches both in F-measure and MAP.
  The source code and output can be accessed in \url{https://github.com/WeiDongjunGabriel/ESA}.
\section{Future Work}\label{F}
  In future work, we expect to try various deep learning methods,
  and design several more powerful and effective neural networks.
  Specifically,
  we may improve our work in the following ways: 
  (1) extend the scale of training set to better train our models;
  (2) instead of employing transE model to tackle the UNK problem, 
  we plan to analyze RDF triples in more fine-grained aspects.

%

%
\bibliographystyle{plain}
\bibliography{reference}

%
\appendix

\end{document}